\documentclass[11pt]{article}

\usepackage[preprint]{acl}

\usepackage{bbm}
\usepackage{booktabs}
\usepackage{hyperref}
\usepackage{url}
\usepackage{xcolor}         
\usepackage{algorithm}
\usepackage{amsmath}
\usepackage{amssymb}
\usepackage{algpseudocode}

\usepackage{multirow}
\usepackage{bbm}
\definecolor{Orange}{RGB}{255,127,0}
\definecolor{MidnightBlue}{RGB}{25,25,112}
\usepackage{microtype}
\usepackage{graphicx}
\usepackage[table]{xcolor}

\usepackage{times}
\usepackage{latexsym}
\usepackage{amsmath}
\usepackage{amsfonts}
\usepackage[T1]{fontenc}

\usepackage[utf8]{inputenc}

\usepackage{microtype}

\usepackage{inconsolata}

\usepackage{graphicx}

\title{Uncertainty-Aware Hybrid Retrieval for Long-Document RAG}

\author{Hoin Jung, Xiaoqian Wang \\
  Elmore Family School of Electrical and Computer Engineering\\
  Purdue University\\
  West Lafayette, IN 47907 \\
  \texttt{\{jung414, joywang\}@purdue.edu} \\}

\begin{document}
\maketitle
\begin{abstract}
Retrieval augmented generation (RAG) depends critically on the quality and granularity of retrieved evidence. Large retrieval units preserve context but often introduce irrelevant content, which can dilute answer bearing evidence and worsen long context utilization. Fine-grained units are more compact, but they may be difficult to retrieve reliably because short chunks can lack semantic, lexical, or bridging cues needed to match the query. We propose \textbf{Uncertainty-aware Multi-Granularity RAG \textsc{(UMG-RAG)}}, a training-free hybrid retrieval framework that treats chunk granularity as query-specific reliability estimation. Instead of training a new retriever or modifying the generator, UMG-RAG uses existing dense and sparse retrievers as complementary experts across multiple chunk granularities. For each query, it converts each expert-granularity score list into an evidence distribution, estimates reliability from distribution entropy, and fuses candidates according to query-specific semantic, lexical, and granularity confidence. We further introduce \textbf{\textsc{UMGP-RAG}}, a parent promotion variant that uses fine-grained hits to locate relevant evidence while returning broader non-redundant parent chunks for local coherence. Experiments on question answering benchmarks show that uncertainty-aware fusion and parent promotion improve generation quality while maintaining a lightweight, plug-and-play retrieval pipeline.
\end{abstract}

\section{Introduction}

Retrieval augmented generation (RAG) \citep{lewis2020retrieval} has become a standard framework for grounding large language models in external knowledge. In a typical pipeline, a retriever first selects relevant text units from a database, and a generator then produces an answer conditioned on the retrieved context. Although this framework reduces the need for fully parametric knowledge storage, its effectiveness depends heavily on the granularity and precision of retrieval \citep{zhang2025fine}. When retrieved units are too large, the generator receives a mixture of relevant and irrelevant content. This can push the model into long context regimes where useful evidence may be diluted or placed in positions that are poorly utilized by the reader, often called the lost-in-the-middle phenomenon \citep{liu2023lost}. When retrieved units are too small, such as isolated sentences, the retrieved evidence may be compact but difficult to recover reliably, because short units may lack semantic cues, entity aliases, or bridging context needed for matching the query. We frame this tension as a retrieval-stage granularity tradeoff between coarse units that preserve context but introduce noise and fine-grained units that provide compact evidence but risk being missed or fragmented.
\begin{figure*}[t]
  \centering
  \includegraphics[width=\linewidth]{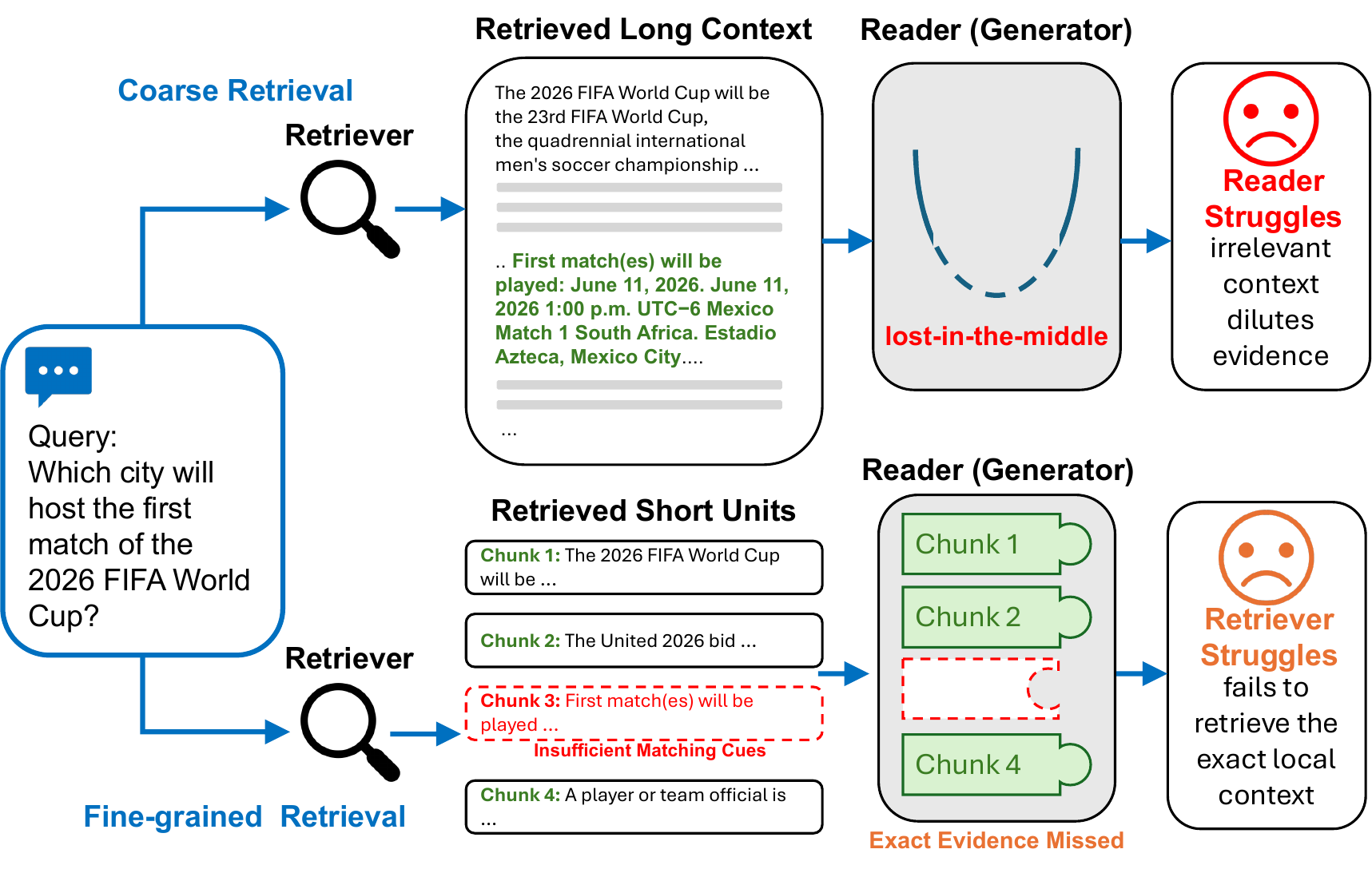}
  \caption{\textbf{Motivation of retrieval granularity tradeoff in RAG.}
  Coarse retrieval units preserve document-level context but may introduce irrelevant content that dilutes answer-bearing evidence. Fine-grained units are more compact, but they may lack sufficient semantic or lexical cues, causing the retriever to miss the exact local evidence.}
  \label{fig:concept1}
  \vspace{-4mm}
\end{figure*}

Figure~\ref{fig:concept1} illustrates this tradeoff. Coarse retrieval can return a long context that contains the answer but also includes substantial irrelevant information, making the generator less likely to use the answer-bearing evidence and potentially reintroducing positional bias in long-context reading. Fine-grained retrieval can avoid this noise, but the exact local evidence may be missed when the query requires surrounding information that is not present in the small unit. This suggests that neither coarse nor fine retrieval units are universally preferable. Instead, an effective retriever should be able to exploit both: fine-grained units are useful when they provide high-confidence evidence, while coarser units are useful when local context is needed to make the evidence retrievable or coherent.

Existing solutions often address only one side of this tradeoff. Long-context interventions calibrate positional attention \citep{hsieh2024found,jung2026cost}, adjust positional encodings \citep{zhang2024found}, reorder documents \citep{wang2025eliminating}, or compress prompts \citep{jiang2024longllmlingua} to reduce the effect of lost-in-the-middle. These methods are valuable, but they generally assume that the relevant evidence has already been retrieved and placed somewhere in the prompt. As a result, they do not directly resolve the retrieval-stage decision of whether the system should retrieve a compact local chunk, a broader parent context, or a combination of both. Advanced RAG systems address fragmentation by constructing knowledge graphs \citep{guo2024lightrag}, retrieving at multiple abstraction levels \citep{zheng2025multiple}, or inferring sentence-level relations \citep{tao-etal-2025-saki}. However, these methods often require graph construction, abstraction-level indexing, or post-retrieval sentence linking, and they do not explicitly estimate query-specific reliability across retrieval sources and granularities.

In this work, we propose Uncertainty-aware Multi-Granularity RAG (\textsc{UMG-RAG}), a training-free hybrid retrieval framework for compact long-document RAG. Instead of training a new retriever or modifying the generator, \textsc{UMG-RAG} uses existing dense and sparse retrievers as complementary experts. Dense retrieval captures semantic similarity, while sparse retrieval captures lexical and entity-level matching. Each expert retrieves candidates across multiple chunk granularities, and the method estimates the confidence of each expert--granularity pair from the sharpness of its candidate score distribution. A peaked distribution indicates that the expert has a clear retrieval preference for the query, while a flat distribution indicates higher retrieval uncertainty. \textsc{UMG-RAG} uses this confidence to fuse dense and sparse evidence into a single ranked list, as shown in Figure~\ref{fig:concept2}.

We further introduce \textsc{UMGP-RAG}, a parent-promotion extension that improves local coherence without giving up fine-grained retrieval precision. When a highly ranked chunk comes from a fine-grained unit, \textsc{UMGP-RAG} promotes it to a broader parent chunk and removes redundant overlaps. This design allows fine-grained retrieval to act as a precise locator of relevant regions, while the final context returned to the generator remains locally coherent. Importantly, the framework is training-free and can be applied on top of existing dense and sparse retrieval backbones.

Our contributions are as follows:
\vspace{-1mm}
\begin{itemize}
    \item We formalize a retrieval granularity tradeoff in long-document RAG, where coarse units improve contextual coverage but add distracting content, while fine-grained units are compact but harder to retrieve reliably.
    \vspace{-1mm}
    \item We propose \textsc{UMG-RAG}, a training-free uncertainty-aware hybrid retrieval method that estimates query-specific reliability for each expert and granularity pair, and further introduce \textsc{UMGP-RAG}, a parent-promotion extension that converts fine-grained hits into broader non-redundant contexts for generation.
    \vspace{-1mm}
    \item We evaluate the proposed methods on long-document RAG benchmarks with multiple dense retrievers and generators, showing that uncertainty-aware fusion and parent promotion improve generation quality while keeping the retrieval pipeline lightweight.
\end{itemize}
\section{Related Work}

\subsection{Long Context Interventions for Lost in the Middle}

Prior work has shown that language models often fail to use relevant evidence when it appears in the middle of a long prompt \citep{liu2024lost}. Recent methods address this issue mainly at the reader or prompt level. FITM \citep{hsieh2024found} calibrates positional attention bias using dummy documents. MS-PoE \citep{zhang2024found} modifies rotary positional encodings to improve long-context utilization. PINE \citep{wang2025eliminating} reassigns document positions based on content importance. LongLLMLingua \citep{jiang2024longllmlingua} compresses long prompts through question-aware pruning and reordering. These approaches are complementary to ours, but they generally assume that relevant evidence has already been retrieved. In contrast, we intervene earlier in the RAG pipeline by constructing a compact, high-signal retrieval context before generation.

\subsection{Retrieval Granularity and Hybrid Retrieval in RAG}

Another line of work modifies the retrieval unit or retrieval structure. LongRAG \citep{jiang2024longrag} increases the retrieval unit size to preserve document-level context, while LightRAG \citep{guo2024lightrag} constructs a knowledge graph to retrieve connected entities and relations. MAL-RAG \citep{zheng2025multiple} retrieves from multiple abstraction levels, including summaries, paragraphs, and sentences. SAKI-RAG \citep{tao-etal-2025-saki} uses sentence-level attention knowledge to recover logically connected evidence. These methods highlight the importance of retrieval granularity, but they often rely on predefined retrieval structures, graph construction, or additional evidence-connection mechanisms.

Hybrid dense-sparse retrieval is also widely used in RAG because dense retrievers capture semantic similarity while sparse retrievers are effective for lexical and entity-level matching. Prior RAG systems such as HyPA-RAG \citep{kalra2024hypa} combine dense and sparse retrieval, often using Reciprocal Rank Fusion as a simple fusion rule \citep{cormack2009reciprocal}. However, fixed fusion does not account for query-specific reliability. A lexical query may favor sparse evidence, while a paraphrastic query may favor dense evidence, and both may require different chunk granularities. Our method differs by estimating confidence for each expert-granularity pair from its score distribution, allowing dense-sparse fusion and retrieval granularity to adapt per query without training a new retriever or modifying the generator.
\begin{figure*}[t]
  \centering
  \includegraphics[width=\linewidth]{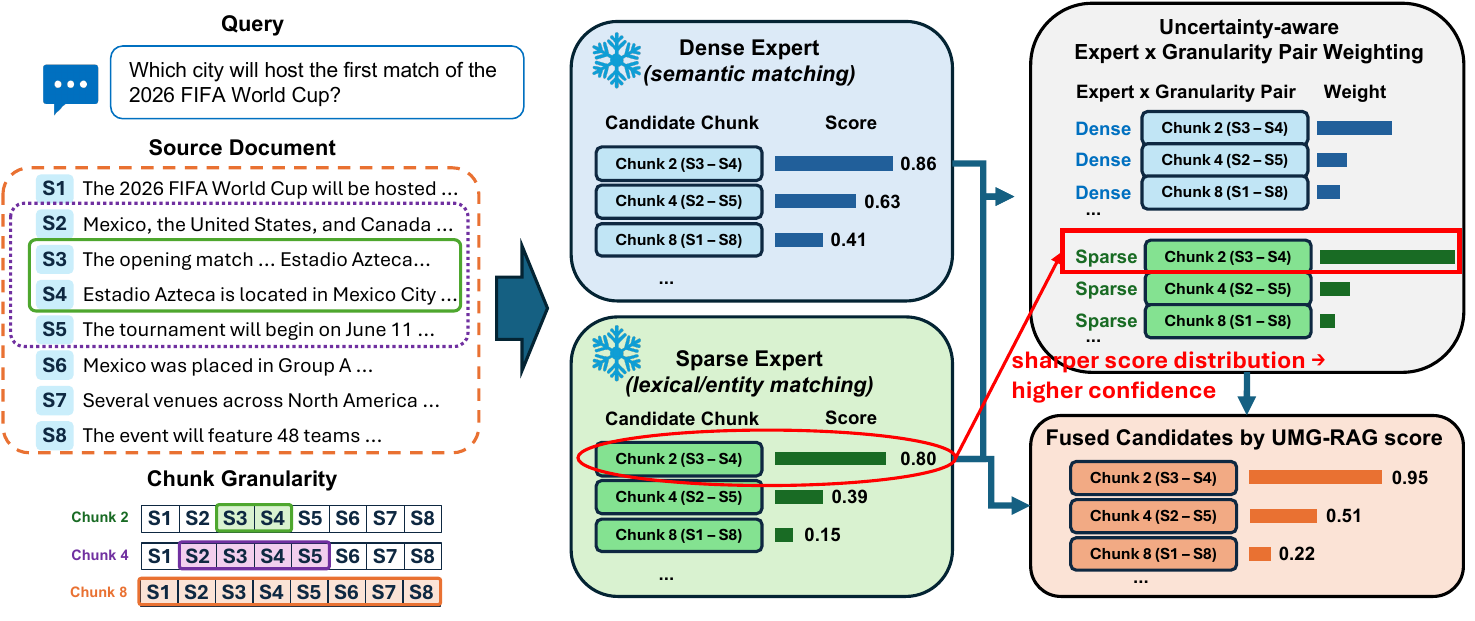}
  \caption{\textbf{Overview of uncertainty-aware multi-granularity RAG.}
  A source document is segmented into overlapping chunks of different granularities. Existing dense and sparse experts retrieve candidates from these granularities. UMG-RAG estimates query-specific confidence for each expert--granularity pair from the sharpness of its score distribution, then fuses the candidates into a single ranked list.}
  \label{fig:concept2}
  \vspace{-4mm}
\end{figure*}

\section{Proposed Method}

\subsection{Overview and Problem Setup}

We propose an uncertainty-aware hybrid retrieval framework for retrieval-augmented generation. Unlike training-based methods, our framework does not introduce a new retriever, train an adapter, or modify the generator. Instead, it uses existing pretrained dense and sparse retrievers as complementary retrieval experts and adaptively selects evidence across multiple chunk granularities without any training.

As motivated in Figure~\ref{fig:concept1}, RAG faces a retrieval granularity tradeoff. Coarse units preserve local context but may introduce distractors, while fine-grained units are compact but can be harder to retrieve reliably. We address this tradeoff by retrieving from multiple granularities and estimating, for each query, which retrieval source and granularity are reliable.

Let $\mathcal{D}=\{d_1,\ldots,d_N\}$ denote a document corpus and let $q$ denote an input query. Each document is segmented into overlapping sentence-level chunks with multiple granularities. We define the granularity set as $
\mathcal{G}=\{2,4,8,16,32\},$
where $g \in \mathcal{G}$ denotes the number of sentences in a retrieval unit. Let $\mathcal{U}_g$ be the set of chunks generated at granularity $g$, and let $
\mathcal{U}=\bigcup_{g \in \mathcal{G}}\mathcal{U}_g
$
be the complete multi-granularity retrieval space.

We use two retrieval experts, 
\[
\mathcal{E}=\{\mathrm{dense\_retriever},\mathrm{sparse\_retriever}\}.
\]
The dense expert captures semantic similarity, while the sparse expert captures lexical and entity-level matching. The framework is model-agnostic and only requires each expert to score a query-chunk pair.

For each expert $e \in \mathcal{E}$ and granularity $g \in \mathcal{G}$, the retriever assigns a raw relevance score
\[
s_{e,g}(q,u) \in \mathbb{R},
\quad
u \in \mathcal{U}_g.
\]
A central challenge is that these scores are not directly comparable. Dense scores, sparse scores, and scores from different chunk granularities may have different scales and distributions. Therefore, using a fixed interpolation weight between dense and sparse retrieval is suboptimal. A fixed weight assumes that the same retriever and granularity are equally reliable for every query, although lexical, entity-centric, and paraphrastic queries may require different retrieval behavior.

Therefore, instead of combining raw scores with a fixed dense--sparse weight, we estimate the query-specific confidence of each expert--granularity pair. This confidence is derived from the uncertainty of its candidate score distribution and determines how much that retrieval source contributes to the final context.

\subsection{Uncertainty-Aware Multi-Granularity RAG}

We now describe \textsc{UMG-RAG}. As shown in Figure~\ref{fig:concept2}, the method retrieves candidates from each expert-granularity pair, converts their scores into normalized evidence distributions, estimates uncertainty using entropy, and fuses candidates into a single ranked list.

For a query $q$, each expert--granularity pair $(e,g)$ first retrieves an intermediate candidate set
\[
\mathcal{C}_{e,g}(q)
=
\mathrm{TopM}_{u \in \mathcal{U}_g}
s_{e,g}(q,u),
\]
where $M$ is the number of candidates retained from each expert--granularity pair before fusion. In our implementation, we set $M=100$. This value is intentionally larger than the final retrieval size $K$ so that the fusion stage can compare a sufficiently diverse set of candidates from both dense and sparse experts across granularities. At the same time, $M=100$ keeps the intermediate pool computationally manageable, since the final context is selected only after uncertainty-aware fusion.

To compare score distributions across experts and granularities, we normalize the raw scores within each candidate set:
\[
\tilde{s}_{e,g}(q,u)
=
\frac{
s_{e,g}(q,u)-\mu_{e,g}(q)
}{
\sigma_{e,g}(q)+\epsilon
},
\]
where $\mu_{e,g}(q)$ and $\sigma_{e,g}(q)$ are the mean and standard deviation of the scores in $\mathcal{C}_{e,g}(q)$, and $\epsilon$ is a small constant for numerical stability.

We then convert the normalized scores into a score-induced evidence distribution:
\begin{equation*}
\small
p_{e,g}(u \mid q,\mathcal{C}_{e,g})  
=
\frac{
\exp(\tilde{s}_{e,g}(q,u))
}{
\sum_{v \in \mathcal{C}_{e,g}(q)}
\exp(\tilde{s}_{e,g}(q,v))
}.
\end{equation*}
This distribution is defined only over the retrieved candidate set. It should not be interpreted as a calibrated probability over the full corpus. Instead, it represents how expert $e$ at granularity $g$ concentrates its evidence on candidate $u$ relative to its other candidates.

This distribution provides a query-specific reliability signal. If an expert--granularity pair assigns most probability mass to a small number of candidates, it has a clear retrieval preference for the current query. If the distribution is nearly uniform, the expert--granularity pair cannot clearly distinguish among its candidates. We quantify this ambiguity using normalized entropy:
\begin{equation*}
\resizebox{.97\linewidth}{!}{$
H_{e,g}(q)
=
-
\frac{
\sum_{u \in \mathcal{C}_{e,g}(q)}
p_{e,g}(u \mid q,\mathcal{C}_{e,g})
\log
p_{e,g}(u \mid q,\mathcal{C}_{e,g})
}{
\log |\mathcal{C}_{e,g}(q)|
}.$}
\end{equation*}
The entropy value lies in $[0,1]$. Low entropy indicates concentrated evidence and therefore low retrieval uncertainty. High entropy indicates diffuse evidence and therefore high retrieval uncertainty.

We convert uncertainty into confidence:
\[
c_{e,g}(q)=1-H_{e,g}(q).
\]
The confidence values are then normalized across all experts and granularities:
\[
w_{e,g}(q)
=
\frac{
c_{e,g}(q)
}{
\sum_{e' \in \mathcal{E}}
\sum_{g' \in \mathcal{G}}
c_{e',g'}(q)
+
\epsilon
}.
\]
This weighting makes the retrieval process uncertainty-aware by assigning larger weights to expert--granularity pairs with lower entropy and smaller weights to pairs with higher entropy.

The final evidence probability of a chunk is defined as the confidence-weighted mixture of candidate distributions:
\begin{equation*}
\resizebox{.97\linewidth}{!}{$
P(u \mid q)
=
\sum_{e \in \mathcal{E}}
\sum_{g \in \mathcal{G}}
w_{e,g}(q)
p_{e,g}(u \mid q,\mathcal{C}_{e,g})
\mathbb{I}
[
u \in \mathcal{C}_{e,g}(q)
].$}
\end{equation*}
Here, $P(u \mid q)$ measures the evidence mass assigned to chunk $u$ after accounting for both retriever confidence and granularity confidence.


Because the generator should receive compact context, we rank chunks by evidence utility:
\[
R(u \mid q)=\frac{P(u \mid q)}{\sqrt{\ell(u)}}.
\]
Here, $\ell(u)$ is the approximate token length of $u$. The square-root penalty mildly favors compact chunks while still allowing longer chunks when they receive strong evidence support.

The final retrieval context is constructed by selecting the top $K$ chunks according to $R(u \mid q)$:
\[
\mathcal{S}_{K}(q)
=
\mathrm{TopK}_{u \in \mathcal{U}_{\mathrm{cand}}}
R(u \mid q), 
\]
$
\mathcal{U}_{\mathrm{cand}}
=
\bigcup_{e \in \mathcal{E}}
\bigcup_{g \in \mathcal{G}}
\mathcal{C}_{e,g}(q)
$ is the fused candidate pool. The selected chunks in $\mathcal{S}_{K}(q)$ are concatenated and used as the retrieval context for the generator. Following common top-$K$ RAG evaluation practice, we set $K=5$ in the main experiments and use the same value across all retrieval methods for a fair comparison.

\subsection{Parent Promotion with Overlap-Aware Deduplication}

We further extend \textsc{UMG-RAG} with parent promotion, denoted \textsc{UMGP-RAG}. The goal is to use fine-grained chunks as precise retrieval signals while returning a broader local unit to the generator.

Let $\pi_8(u)$ denote the parent chunk of $u$ at granularity $8$. In our implementation, retrieved chunks from granularity $2$ or $4$ are promoted to their corresponding parent chunk:
\[
\rho(u)
=
\begin{cases}
\pi_8(u), & u \in \mathcal{U}_2 \cup \mathcal{U}_4, \\
u, & \text{otherwise},
\end{cases}
\]
where $\rho(u)$ is the unit returned after parent promotion. Thus, fine-grained chunks serve as retrieval locators, while the returned parent chunks provide the neighboring sentences needed for local coherence.

After applying $\rho(\cdot)$, multiple fine-grained candidates may map to the same parent chunk. We merge such duplicates using the same bounded evidence aggregation used in the implementation. Let $\mathcal{M}(v)=\{u \in \mathcal{U}_{\mathrm{cand}}:\rho(u)=v\}$ be the set of candidates mapped to promoted chunk $v$. We define
\[
A(v \mid q)
=
1-
\prod_{u \in \mathcal{M}(v)}
\left(1-P(u \mid q)\right).
\]
This aggregation increases the evidence score of a parent chunk when multiple retrieved candidates support it, while keeping the score bounded in $[0,1]$. Promoted candidates are then ranked by
\[
R_{\mathrm{P}}(v \mid q)
=
\frac{A(v \mid q)}{\sqrt{\ell(v)}}.
\]
To avoid returning redundant contexts, we remove promoted candidates that substantially overlap with already selected chunks. Let $\mathrm{span}(v)$ denote the sentence span covered by candidate chunk $v$. For a candidate $v$ and a previously selected chunk $s$, we define span overlap as
\[
\mathrm{overlap}(v,s)
=
\frac{
|\mathrm{span}(v)\cap \mathrm{span}(s)|
}{
\min(|\mathrm{span}(v)|,|\mathrm{span}(s)|)
}.
\]
A candidate is skipped if at least 75\% of the shorter span overlaps with any previously selected chunk. This rule removes near-duplicate or nested contexts while allowing partially overlapping chunks to remain when they may provide complementary evidence.

The final context is constructed by selecting the top $K$ non-redundant promoted chunks:
\[
\mathcal{S}^{\mathrm{P}}_{K}(q)
=
\mathrm{TopKDedup}_{v \in \{\rho(u):u \in \mathcal{U}_{\mathrm{cand}}\}}
R_{\mathrm{P}}(v \mid q),
\]
where $\mathcal{U}_{\mathrm{cand}}$ is the fused candidate pool produced by \textsc{UMG-RAG}. The selected promoted chunks are concatenated and used as the final retrieval context for the generator. The detailed algorithm is introduced in Algorithm \ref{alg:umgr_p} in Appendix \ref{sec:UMGR-P}.

\begin{table*}[t]
\fontsize{9.5pt}{11pt}\selectfont
\centering
\begin{tabular}{c|c|c|cc|cc}
\toprule
\multirow{2}{*}{Retriever} & \multirow{2}{*}{Method} & \multirow{2}{*}{\begin{tabular}[c]{@{}c@{}}Retrieval\\  AR@5 $\uparrow$\end{tabular}} & \multicolumn{2}{c|}{Qwen2.5-3B-Instruct} & \multicolumn{2}{c}{Llama-3.2-3B-Instruct} \\ \cmidrule{4-7}
                           &                         &                                 & F1 $\uparrow$         & AR $\uparrow$        & F1 $\uparrow$          & AR  $\uparrow$        \\ \midrule
SPLADEv3                   &         RAG             & 0.7994                          & 0.4745            &     0.4529       &         0.4923    & 0.4741           \\ \midrule
\multirow{10}{*}{BERT}      &        RAG              &  \multirow{6}{*}{0.4477}        & 0.2270            &  0.2013          & 0.2414            &  0.2178           \\
                           &     Summarized RAG      &                                 &  0.1704           &   0.1491         &  0.1972           &  0.1703          \\
                           &     LongLLMLingua       &                                 &   0.2304          &  0.2080          &   0.2573          & 0.2429           \\
                           &     MS-PoE              &&0.2239&0.1974&0.2435&0.2192\\
                           &     REPLUG              &&0.1373&0.1174&0.2254&0.1896\\
                           &     PINE                &&0.1915&0.1843&0.2084&0.2097\\\cmidrule{2-7}
                           &     LongRAG             &  0.4441                               &  0.2093           &   0.1851         & 0.2318            & 0.2094           \\
                           &     Hybrid              &\underline{0.7011}&0.4265&0.4018&0.4504&0.4293\\
                           &  \cellcolor{gray!15}   \textbf{UMG-RAG (Ours)}      &\cellcolor{gray!15} 0.6852&\cellcolor{gray!15}\underline{0.4350}&\cellcolor{gray!15}\underline{0.4110}&\cellcolor{gray!15}\underline{0.4514}&\cellcolor{gray!15}\underline{0.4304}\\
                           &\cellcolor{gray!15}     \textbf{UMGP-RAG (Ours)}     &\cellcolor{gray!15}\textbf{0.7797}&\cellcolor{gray!15}\textbf{0.4572}&\cellcolor{gray!15}\textbf{0.4307}&\cellcolor{gray!15}\textbf{0.4811}&\cellcolor{gray!15}\textbf{0.4533}\\ \midrule
\multirow{10}{*}{BGE-M3}    &        RAG              &  \multirow{6}{*}{0.8562}        &  0.4407     &  0.3994           &      0.4941       &     0.4593       \\
                           &     Summarized RAG      &&0.3243&0.2880&0.3520&0.3127\\
                           &     LongLLMLingua       &&0.4581&0.4184&0.4631&0.4462\\
                           &     MS-PoE              
                           &&0.4432&0.4015&0.5005&0.4709\\
                           &     REPLUG              &&0.4701&0.4216&0.4962&0.4459\\
                           &     PINE                &&0.4547&0.4420&0.4452&0.4603\\\cmidrule{2-7}
                           &     LongRAG             &\textbf{0.9101}&0.4598&0.4219&\underline{0.5113}&\textbf{0.4910}\\
                           &     Hybrid              &0.8241&\underline{0.4927}&\underline{0.4727}&0.4986&0.4766\\
                           &    \cellcolor{gray!15} \textbf{UMG-RAG (Ours)}      &\cellcolor{gray!15}0.8023&\cellcolor{gray!15}{0.4809}&\cellcolor{gray!15}{0.4593}&\cellcolor{gray!15}0.4873&\cellcolor{gray!15}0.4716\\
                           &    \cellcolor{gray!15} \textbf{UMGP-RAG (Ours)}     &\cellcolor{gray!15}\underline{0.8759}&\cellcolor{gray!15}\textbf{0.5052}&\cellcolor{gray!15}\textbf{0.4794}&\cellcolor{gray!15}\textbf{0.5128}&\cellcolor{gray!15}\underline{0.4878}\\ \midrule
\multirow{10}{*}{\begin{tabular}[c]{@{}c@{}}Qwen3-\\ Embedding-4B\end{tabular}} &      RAG              &  \multirow{6}{*}{0.8601}        &     0.4427&0.3990&0.4939&0.4621\\
                           &     Summarized RAG      &&0.3333&0.2915&0.3537&0.3162\\
                           &     LongLLMLingua       &&0.4459&0.4057&0.4561&0.4346\\
                           &     MS-PoE              
                           &&0.4481&0.4064&0.5046&0.4720\\
                           &     REPLUG              &&0.4796&0.4367&0.5120&0.4642\\
                           &     PINE                &&0.4628&0.4466&0.4291&0.4438\\\cmidrule{2-7}
                           &     LongRAG             &\textbf{0.9253}&0.4744&0.4329&\underline{0.5145}&\underline{0.4949}\\
                           &     Hybrid              &0.8114&0.4803&\underline{0.4621}&0.4916&0.4741\\
                           &  \cellcolor{gray!15}   \textbf{UMG-RAG (Ours)}      &\cellcolor{gray!15}0.7949&\cellcolor{gray!15}\underline{0.4807}&\cellcolor{gray!15}0.4582&\cellcolor{gray!15}0.4974&\cellcolor{gray!15}0.4790\\
                           &  \cellcolor{gray!15}   \textbf{UMGP-RAG (Ours)}     &\cellcolor{gray!15}\underline{0.8745}&\cellcolor{gray!15}\textbf{0.4961}&\cellcolor{gray!15}\textbf{0.4748}&\cellcolor{gray!15}\textbf{0.5220}&\cellcolor{gray!15}\textbf{0.4984}\\ \bottomrule
                           
\end{tabular}
\vspace{-1mm}
\caption{
\textbf{Results on the Natural Questions dataset.}
The best retrieval and generation scores within each retriever block are highlighted in \textbf{bold}, and the second best scores are \underline{underlined}.
}
\label{tab:nq_result}
\vspace{-1.5mm}
\end{table*}

\begin{table*}[t]
\fontsize{9.5pt}{11pt}\selectfont
\centering
\begin{tabular}{c|c|c|cc|cc}
\toprule
\multirow{2}{*}{Retriever} & \multirow{2}{*}{Method} & \multirow{2}{*}{\begin{tabular}[c]{@{}c@{}}Retrieval\\  AR@5 $\uparrow$\end{tabular}} & \multicolumn{2}{c|}{Qwen2.5-3B-Instruct} & \multicolumn{2}{c}{Llama-3.2-3B-Instruct} \\ \cmidrule{4-7}
                           &                         &                                 & F1 $\uparrow$         & AR $\uparrow$        & F1 $\uparrow$          & AR  $\uparrow$        \\ \midrule
SPLADEv3                   &         RAG             &0.6043&0.4613&0.3850&0.4798&0.3966\\ \midrule
\multirow{10}{*}{BERT}      &        RAG              &\multirow{6}{*}{0.3638}&0.2599&0.2087&0.2803&0.2279\\
                           &     Summarized RAG      &&0.2275&0.1864&0.2469&0.2025\\
                           &     LongLLMLingua       &&0.2833&0.2271&0.3021&0.2362\\
                           &     MS-PoE              
                           &&0.2612&0.2112&0.2860&0.2339\\
                           &     REPLUG              &&0.1798&0.1466&0.2560&0.1984\\
                           &     PINE                &&0.2424&0.2039&0.2557&0.2180\\\cmidrule{2-7}
                           &     LongRAG             &0.3230&0.2299&0.1861&0.2625&0.2104\\
                           &     Hybrid              &\underline{0.5207}&0.4081&0.3311&\underline{0.4245}&\underline{0.3479}\\
                           &    \cellcolor{gray!15} \textbf{UMG-RAG (Ours)}      &\cellcolor{gray!15}0.5134&\cellcolor{gray!15}\underline{0.4163}&\cellcolor{gray!15}\underline{0.3416}&\cellcolor{gray!15}0.4223&\cellcolor{gray!15}0.3470\\
                           &     \cellcolor{gray!15}\textbf{UMGP-RAG (Ours)}     &\cellcolor{gray!15}\textbf{0.5830}&\cellcolor{gray!15}\textbf{0.4453}&\cellcolor{gray!15}\textbf{0.3645}&\cellcolor{gray!15}\textbf{0.4460}&\cellcolor{gray!15}\textbf{0.3655}\\ \midrule
\multirow{10}{*}{BGE-M3}    &        RAG              &\multirow{6}{*}{\textbf{0.7811}}&0.4613&0.3798&0.4913&0.4048\\
                           &     Summarized RAG      &&0.3856&0.3161&0.4109&0.3366\\
                           &     LongLLMLingua       &&0.4706&0.3807&0.4668&0.3771\\
                           &     MS-PoE              
                           &&0.4622&0.3805&\underline{0.5001}&0.4123\\
                           &     REPLUG              &&0.3941&0.3150&0.4287&0.3362\\
                           &     PINE                &&0.4442&0.3787&0.4409&0.3941\\\cmidrule{2-7}
                           &     LongRAG             &\underline{0.7457}&0.4658&0.3845&0.4799&0.3984\\
                           &     Hybrid              &0.6471&\underline{0.4775}&\underline{0.3975}&{0.4986}&\underline{0.4175}\\
                           &    \cellcolor{gray!15} \textbf{UMG-RAG (Ours)}      &\cellcolor{gray!15}0.6273&\cellcolor{gray!15}0.4749&\cellcolor{gray!15}0.3957&\cellcolor{gray!15}0.4827&\cellcolor{gray!15}0.4018\\
                           &     \cellcolor{gray!15}\textbf{UMGP-RAG (Ours)}     &\cellcolor{gray!15}0.7002&\cellcolor{gray!15}\textbf{0.5034}&\cellcolor{gray!15}\textbf{0.4216}&\cellcolor{gray!15}\textbf{0.5133}&\cellcolor{gray!15}\textbf{0.4275}\\ \midrule
\multirow{10}{*}{\begin{tabular}[c]{@{}c@{}}Qwen3-\\ Embedding-4B\end{tabular}} &      RAG              &  \multirow{6}{*}{\textbf{0.7971}}        &0.4883&0.4041&0.4991&0.4132

\\
                           &     Summarized RAG      &&0.3976&0.3261&0.4315&0.3546\\
                           &     LongLLMLingua       &&0.4603&0.3693&0.4644&0.3738\\
                           &     MS-PoE              
                           &&\textbf{0.4948}&\textbf{0.4073}&\textbf{0.5172}&\textbf{0.4295}\\
                           &     REPLUG              &&0.4075&0.3252&0.4343&0.3411\\
                           &     PINE                &&0.4621&0.3980&0.4463&0.4027\\\cmidrule{2-7}
                           &     LongRAG             &\underline{0.7280}&0.4524&0.3748&0.4765&0.3961\\
                           &     Hybrid              &0.6091&0.4626&0.3837&0.4709&0.3904\\
                           &    \cellcolor{gray!15} \textbf{UMG-RAG (Ours)}      &\cellcolor{gray!15}0.5845&\cellcolor{gray!15}0.4622&\cellcolor{gray!15}0.3820&\cellcolor{gray!15}0.4752&\cellcolor{gray!15}0.3929\\
                           &     \cellcolor{gray!15}\textbf{UMGP-RAG (Ours)}     &\cellcolor{gray!15}0.6761&\cellcolor{gray!15}\underline{0.4919}&\cellcolor{gray!15}\underline{0.4071}&\cellcolor{gray!15}\underline{0.5068}&\cellcolor{gray!15}\underline{0.4200}\\ \bottomrule
                           
\end{tabular}
\caption{
\textbf{Results on the HotPotQA dataset.}
The best retrieval and generation scores within each retriever block are highlighted in \textbf{bold}, and the second best scores are \underline{underlined}.
}
\label{tab:hotpotqa_result}
\end{table*}
\section{Implementation Details}
\vspace{-1mm}
\subsection{Models and Datasets}

We evaluate our RAG pipeline on Natural Questions (NQ) \citep{kwiatkowski2019natural} and HotPotQA \citep{yang-etal-2018-hotpotqa}, following prior long-document RAG evaluation settings \citep{li2024retrieval,petroni2020kilt}. NQ evaluates factual question answering over external knowledge, while HotPotQA requires more compositional evidence retrieval and provides a complementary testbed for multi-hop reasoning.

For each dataset, we construct a task-specific retrieval corpus from the KILT \citep{petroni2020kilt} validation split. The corpus contains all Wikipedia documents appearing in the gold provenance annotations, plus 10,000 randomly sampled distractor Wikipedia documents that do not overlap with the positive set. This results in 14,274 documents for NQ, consisting of 4,274 positive documents and 10,000 distractors, and 20,519 documents for HotPotQA, consisting of 10,519 positive documents and 10,000 distractors. We use the same corpus for all retrieval methods to ensure that performance differences come from the retrieval strategy rather than the candidate document pool.

For generators, we use \textsf{Qwen2.5-3B-Instruct} \citep{qwen2} and \textsf{Llama-3.2-3B-Instruct} \citep{grattafiori2024llama}. For dense retrievers, we evaluate BERT \citep{devlin2019bert}, BGE-M3 \citep{bge-m3}, and \textsf{Qwen3-Embedding-4B} \citep{qwen3embedding}. For sparse retrieval, we use SPLADEv3 \citep{lassance2024splade}. All methods are evaluated without task specific training or finetuning, so performance differences reflect retrieval and context construction rather than updates to the retriever or generator parameters.
\subsection{Evaluation Metrics}

We evaluate both retrieval quality and generation quality. For retrieval, we report answer recall at five retrieved chunks (AR@5). A retrieval is counted as successful if the normalized gold answer appears in any of the top five retrieved contexts.

For generation, we report token-level F1 and answer recall (AR). F1 measures token overlap between the generated answer and the gold answer after normalization. Generation AR measures whether the normalized gold answer appears in the generated response. We follow standard question answering normalization by lowercasing text, removing punctuation, and removing articles. Higher values indicate better performance for all metrics.

For readability, the main tables report mean performance. We additionally report standard deviations estimated from 1,000 bootstrap resamples in Appendix~\ref{sec:statistical_robustness}.
\subsection{Baselines}

We compare our methods against standard and long context RAG baselines. The standard RAG baseline retrieves from a fixed retrieval unit and passes the retrieved text directly to the generator. Summarized RAG compresses retrieved documents using the Falcon summarization model \citep{falconsai_summarization}. REPLUG \citep{shi2024replug} represents a retrieval-augmented black-box language modeling baseline that combines retrieved documents without updating the generator. LongLLMLingua \citep{jiang2024longllmlingua} represents prompt compression based context selection, while MS-PoE \citep{zhang2024found} and PINE \citep{wang2025eliminating} represent long context reader side interventions for mitigating position bias. LongRAG \citep{jiang2024longrag} uses larger retrieval units to preserve broader context.

We also include a simple hybrid retrieval baseline. Following common hybrid retrieval practice in RAG \citep{kalra2024hypa}, we combine dense and sparse retrieval results using Reciprocal Rank Fusion (RRF) \citep{cormack2009reciprocal}, denoted Hybrid. This baseline tests whether fixed dense-sparse fusion is sufficient. In contrast, our \textsc{UMG-RAG} estimates query specific confidence for each expert and granularity before fusion. \textsc{UMGP-RAG} further promotes fine grained hits to parent chunks and removes redundant overlaps. All comparison methods are used in a plug-and-play setting without additional training, to isolate the effect of retrieval strategy.
\section{Results and Analysis}

Tables~\ref{tab:nq_result} and~\ref{tab:hotpotqa_result} report retrieval and generation performance on NQ and HotPotQA. We report retrieval answer recall at five retrieved chunks (AR@5), and generation performance using F1 and answer recall (AR) with two instruction tuned generators. 
Across most retriever and generator combinations, \textsc{UMGP-RAG} achieves the strongest or highly competitive generation performance.
This result supports the central hypothesis of this work: the best RAG context is not necessarily the longest context or the context with the highest raw retrieval hit rate, but the context that is compact, locally coherent, and aligned with the query. In several cases, standard dense retrieval or LongRAG can achieve competitive retrieval AR@5 because the answer string appears somewhere in a long retrieved passage. However, such passages may contain substantial irrelevant content and can reintroduce positional bias when the answer-bearing span appears in a less favorable location, making it harder for the generator to identify and use the correct evidence. \textsc{UMGP-RAG} instead uses fine-grained retrieval to locate precise evidence and parent promotion to return a coherent local context, reducing distractor content and improving downstream generation.

Although the BERT-based variants do not outperform SPLADE-only, they still improve over other BERT-based RAG baselines. This result is expected because BERT is a weaker dense retriever than recent embedding models, while SPLADE provides strong lexical matching for answer-bearing spans. The goal of \textsc{UMG-RAG} is not to replace a strong sparse retriever, but to adaptively combine dense and sparse evidence when both sources are available. The stronger performance with BGE-M3 and Qwen3-Embedding-4B further suggests that the proposed framework benefits from stronger dense experts.

\vspace{-2mm}
\paragraph{Ablation study.}
The comparison between \textsc{UMG-RAG} and \textsc{UMGP-RAG} further shows the value of parent promotion. \textsc{UMG-RAG} evaluates the effect of uncertainty aware dense-sparse fusion across granularities, while \textsc{UMGP-RAG} adds parent promotion and overlap aware deduplication. The consistent improvement of \textsc{UMGP-RAG} indicates that adaptive fusion is most effective when fine-grained hits are converted into coherent parent contexts. This supports our design choice of using small chunks for retrieval signals and parent chunks for final context construction.
\section{Conclusion}

We introduce \textsc{UMG-RAG}, a training-free uncertainty-aware hybrid retrieval framework for compact long document RAG. The method uses existing dense and sparse retrievers across multiple chunk granularities and estimates query specific confidence from candidate score distributions. We further propose \textsc{UMGP-RAG}, which promotes fine grained hits to parent chunks and removes redundant overlaps to improve local coherence. Experiments on NQ and HotPotQA show that uncertainty-aware fusion and parent promotion improve generation quality while preserving a lightweight pipeline. These results suggest that effective RAG requires retrieving not only answer-containing passages, but compact, coherent contexts that the generator can use reliably.

\section*{Limitations}

\paragraph{Retrieval preprocessing cost.}
UMG-RAG and UMGP-RAG retrieve candidates across multiple granularities using both dense and sparse retrievers. As shown in the computational cost analysis in Appendix~\ref{sec:cost}, this increases retrieval preprocessing cost compared with single-granularity RAG baselines. This overhead is the main cost of the proposed framework. However, the additional computation occurs before generation and is used to construct a more compact context, which can reduce generation time and memory usage.

\paragraph{Dependence on existing retrieval experts.}
The proposed framework is training-free and does not update the dense retriever, sparse retriever, or generator. This makes the method lightweight and easy to apply, but it also means that its performance depends on the quality of the underlying retrieval experts. If both dense and sparse retrievers fail to assign useful scores to answer-bearing chunks, uncertainty-aware fusion cannot fully recover the missing evidence. Thus, UMG-RAG is best understood as an adaptive fusion and context construction framework rather than a replacement for strong retrieval models.

\section*{Ethical Considerations}

This work aims to improve retrieval and context construction for RAG systems. The proposed method does not introduce new training data, does not finetune the retriever or generator, and does not require collecting user information. However, it relies on external corpora and pretrained retrieval and generation models, which may contain factual errors, outdated information, or social biases. Improving retrieval quality may increase the likelihood that a model uses retrieved evidence, but it does not guarantee that the final generated answer is correct, unbiased, or safe. In real-world deployments, corpus curation, privacy protection, and output monitoring remain necessary.

\bibliography{reference}

\appendix
\newpage
\section{Detailed Algorithm of UMGP-RAG}
\label{sec:UMGR-P}

\begin{algorithm}[h]
\small
\caption{\textsc{UMGP-RAG}}
\label{alg:umgr_p}
\begin{algorithmic}[1]
\Require Query $q$, experts $\mathcal{E}$, granularities $\mathcal{G}$, $M=100$, $K=5$, overlap cutoff $0.75$
\Ensure Selected context chunks $\mathcal{S}$

\State $\mathcal{U}_{\mathrm{cand}} \gets \emptyset$

\ForAll{$(e,g) \in \mathcal{E} \times \mathcal{G}$}
    \State $\mathcal{C}_{e,g}(q) \gets \operatorname{TopM}_{u \in \mathcal{U}_g} s_{e,g}(q,u)$
    \State Compute $p_{e,g}(u \mid q,\mathcal{C}_{e,g})$ from normalized scores
    \State Compute $H_{e,g}(q)$ and $c_{e,g}(q)=1-H_{e,g}(q)$
    \State $\mathcal{U}_{\mathrm{cand}} \gets \mathcal{U}_{\mathrm{cand}} \cup \mathcal{C}_{e,g}(q)$
\EndFor

\State Normalize $\{c_{e,g}(q)\}$ to obtain $w_{e,g}(q)$
\State Compute $P(u \mid q)$ and $R(u \mid q)=P(u \mid q)/\sqrt{\ell(u)}$ for all $u \in \mathcal{U}_{\mathrm{cand}}$
\State Sort $\mathcal{U}_{\mathrm{cand}}$ by $R(u \mid q)$ in descending order

\State $\mathcal{U}_{\mathrm{promoted}} \gets \emptyset$, $A(v)\gets 0$

\ForAll{$u$ in ranked $\mathcal{U}_{\mathrm{cand}}$}
    \State $v \gets \pi_8(u)$ if $u \in \mathcal{U}_2 \cup \mathcal{U}_4$, otherwise $v \gets u$
    \State $\mathcal{U}_{\mathrm{promoted}} \gets \mathcal{U}_{\mathrm{promoted}} \cup \{v\}$
    \State $A(v) \gets 1-(1-A(v))(1-P(u \mid q))$
\EndFor

\State Sort $\mathcal{U}_{\mathrm{promoted}}$ by $A(v)$ in descending order
\State $\mathcal{S} \gets \emptyset$

\ForAll{$v$ in ranked $\mathcal{U}_{\mathrm{promoted}}$}
    \If{$|\mathcal{S}| = K$}
        \State \textbf{break}
    \EndIf
    \If{$\mathcal{S}=\emptyset$ or $\max_{s \in \mathcal{S}}\mathrm{overlap}(v,s) < 0.75$}
        \State $\mathcal{S} \gets \mathcal{S} \cup \{v\}$
    \EndIf
\EndFor

\State \Return $\mathcal{S}$
\end{algorithmic}
\end{algorithm}
\begin{table*}[!ht]
\centering
\small
\begin{tabular}{l c c c c}
\toprule
\textbf{Method} & \begin{tabular}[c]{@{}c@{}}\textbf{Retrieval} \\\textbf{Preprocessing Time} \\
 \textbf{(s/question)}\end{tabular} & \begin{tabular}[c]{@{}c@{}}\textbf{Retrieval}\\ \textbf{Preprocessing Mem.} \\
 \textbf{(MiB)}\end{tabular} &\begin{tabular}[c]{@{}c@{}}\textbf{Generation Time} \\
 \textbf{(s/question)}\end{tabular} &\begin{tabular}[c]{@{}c@{}}\textbf{Generation Mem.} \\
 \textbf{(MiB)}\end{tabular} \\
\midrule
Standard RAG & \multirow{7}{*}{0.15}&\multirow{7}{*}{7954}&0.71&7558\\
Summ. RAG &&&0.28&7594\\
LongLLMLingua&&&4.70&24398\\
MS-PoE &&&0.71&7554\\
REPLUG &&&1.01&9452\\
PINE &&&1.03&7558\\
LongRAG &&&0.71&7618\\
\midrule
Hybrid &1.59&8396&0.27&6908\\
\textbf{UMG-RAG}&5.16&8396&0.30&6762\\
\textbf{UMGP-RAG}&5.36&8396&0.33&6716\\
\bottomrule
\end{tabular}
\caption{Computational cost analysis of the retrieval preprocessing and generation phases on the NQ dataset using Qwen3-Embedding-4B and Llama-3.2-3B-Instruct. Time is reported in seconds per question and peak memory in MiB. While the proposed multi-granularity methods require additional time during the retrieval preprocessing step, they achieve the most efficient generation phase with the lowest time and memory requirements.}
\label{tab:generation_cost}
\end{table*}

\section{Computational Cost Analysis}
\label{sec:cost}
Table~\ref{tab:generation_cost} presents a comprehensive computational cost analysis of both the retrieval preprocessing and generation phases across the evaluated methods. This investigation is conducted on the NQ dataset, utilizing the Qwen3-Embedding-4B model for dense retrieval and the Llama-3.2-3B-Instruct generator as detailed in the main paper. The baseline models exhibit a fast retrieval preprocessing time of 0.15 seconds per question. In contrast, our proposed multi-granularity approaches require higher retrieval preprocessing times, peaking at 5.36 seconds for UMGP-RAG due to multi-granualirty. This increased overhead is the main computational cost of the framework, caused by multi-granularity dense and sparse retrieval. However, the resulting compact context reduces generation time and memory usage during the computationally demanding generation phase. By ensuring that only a highly relevant and compact context is fed into the language model, our approaches achieve the lowest generation times (between 0.30 and 0.33 seconds per question) and the smallest peak memory footprints (under 7000 MiB). This demonstrates that the initial investment in thorough database preparation directly translates to substantial speedups and memory reductions when generating the final response.

\section{Statistical Robustness with Bootstrap Resampling}
\label{sec:statistical_robustness}

The main results report mean retrieval and generation performance for readability. To assess statistical robustness, we additionally estimate standard deviations using 1,000 bootstrap resamples over the evaluation examples. Each bootstrap sample is drawn with replacement from the original evaluation set, and all metrics are recomputed on the resampled set. Tables~\ref{tab:nq_result_qwen} and~\ref{tab:nq_result_llama} report the bootstrap results on Natural Questions with Qwen2.5-3B-Instruct and Llama-3.2-3B-Instruct, respectively. Tables~\ref{tab:hotpotqa_result_qwen} and~\ref{tab:hotpotqa_result_llama} report the corresponding results on HotPotQA.

Overall, the bootstrap results show that the main trends are stable. \textsc{UMGP-RAG} consistently improves over \textsc{UMG-RAG}, supporting the benefit of parent promotion and overlap-aware deduplication. The results also show that the strongest retrieval AR@5 does not always correspond to the strongest generation performance. For example, long-context retrieval methods can achieve high answer coverage, but the resulting context may still include distractors or place answer-bearing evidence in less favorable positions for the generator. In contrast, \textsc{UMGP-RAG} tends to produce stronger generation performance by constructing compact and locally coherent contexts.

The standard deviations are small relative to the observed performance differences in most settings, indicating that the improvements are not driven by a small number of examples. These results further support the conclusion that uncertainty-aware fusion and parent promotion provide a robust benefit for long-document RAG.
\begin{table*}[t]
\fontsize{9.5pt}{11pt}\selectfont
\centering
\begin{tabular}{c|c|c|cc}
\toprule
\multirow{2}{*}{Retriever} & \multirow{2}{*}{Method} & \multirow{2}{*}{\begin{tabular}[c]{@{}c@{}}Retrieval\\ AR@5 $\uparrow$\end{tabular}} & \multicolumn{2}{c}{Qwen2.5-3B-Instruct} \\ \cmidrule{4-5}
                           &                         &                                                                 & F1 $\uparrow$         & AR $\uparrow$        \\ \midrule
SPLADEv3                   &         RAG             & 0.7994$\pm$0.0074                                               & 0.4745$\pm$0.0083 & 0.4529$\pm$0.0094 \\ \midrule
\multirow{10}{*}{BERT}     &        RAG              & \multirow{6}{*}{0.4477$\pm$0.0090}                              & 0.2270$\pm$0.0069 & 0.2013$\pm$0.0072 \\
                           &      Summarized RAG     &                                                                 & 0.1704$\pm$0.0064 & 0.1491$\pm$0.0068 \\
                           &      LongLLMLingua      &                                                                 & 0.2304$\pm$0.0071 & 0.2080$\pm$0.0078 \\
                           &      MS-PoE             &                                                                 & 0.2239$\pm$0.0070 & 0.1974$\pm$0.0074 \\
                           &      REPLUG             &                                                                 & 0.1373$\pm$0.0059 & 0.1174$\pm$0.0059 \\
                           &      PINE               &                                                                 & 0.1915$\pm$0.0066 & 0.1843$\pm$0.0072 \\\cmidrule{2-5}
                           &      LongRAG            & 0.4441$\pm$0.0093                                               & 0.2093$\pm$0.0070 & 0.1851$\pm$0.0073 \\
                           &      Hybrid             & \underline{0.7011$\pm$0.0086}                                   & 0.4265$\pm$0.0090 & 0.4018$\pm$0.0093 \\
                           &  \cellcolor{gray!15}   \textbf{UMG-RAG (Ours)}      &\cellcolor{gray!15} 0.6852$\pm$0.0088&\cellcolor{gray!15}\underline{0.4350$\pm$0.0084}&\cellcolor{gray!15}\underline{0.4110$\pm$0.0092}\\
                           &\cellcolor{gray!15}      \textbf{UMGP-RAG (Ours)}     &\cellcolor{gray!15}\textbf{0.7797$\pm$0.0078}&\cellcolor{gray!15}\textbf{0.4572$\pm$0.0087}&\cellcolor{gray!15}\textbf{0.4307$\pm$0.0091}\\ \midrule
\multirow{10}{*}{BGE-M3}   &         RAG             & \multirow{6}{*}{0.8562$\pm$0.0068}                              & 0.4407$\pm$0.0085 & 0.3994$\pm$0.0089 \\
                           &      Summarized RAG     &                                                                 & 0.3243$\pm$0.0080 & 0.2880$\pm$0.0085 \\
                           &      LongLLMLingua      &                                                                 & 0.4581$\pm$0.0083 & 0.4184$\pm$0.0089 \\
                           &      MS-PoE             &                                                                 & 0.4432$\pm$0.0083 & 0.4015$\pm$0.0092 \\
                           &      REPLUG             &                                                                 & 0.4701$\pm$0.0082 & 0.4216$\pm$0.0093 \\
                           &      PINE               &                                                                 & 0.4547$\pm$0.0084 & 0.4420$\pm$0.0092 \\\cmidrule{2-5}
                           &      LongRAG            & \textbf{0.9101$\pm$0.0055}                                      & 0.4598$\pm$0.0083 & 0.4219$\pm$0.0093 \\
                           &      Hybrid             & 0.8241$\pm$0.0071                                               & \underline{0.4927$\pm$0.0085}& \underline{0.4727$\pm$0.0092}\\
                           &    \cellcolor{gray!15} \textbf{UMG-RAG (Ours)}      &\cellcolor{gray!15}0.8023$\pm$0.0074&\cellcolor{gray!15}{0.4809$\pm$0.0081}&\cellcolor{gray!15}{0.4593$\pm$0.0093}\\
                           &    \cellcolor{gray!15} \textbf{UMGP-RAG (Ours)}     &\cellcolor{gray!15}\underline{0.8759$\pm$0.0063}&\cellcolor{gray!15}\textbf{0.5052$\pm$0.0086}&\cellcolor{gray!15}\textbf{0.4794$\pm$0.0092}\\ \midrule
\multirow{10}{*}{\begin{tabular}[c]{@{}c@{}}Qwen3-\\ Embedding-4B\end{tabular}} &       RAG              &  \multirow{6}{*}{0.8601$\pm$0.0067}                             & 0.4427$\pm$0.0084 & 0.3990$\pm$0.0092 \\
                           &      Summarized RAG     &                                                                 & 0.3333$\pm$0.0084 & 0.2915$\pm$0.0086 \\
                           &      LongLLMLingua      &                                                                 & 0.4459$\pm$0.0085 & 0.4057$\pm$0.0089 \\
                           &      MS-PoE             &                                                                 & 0.4481$\pm$0.0083 & 0.4064$\pm$0.0090 \\
                           &      REPLUG             &                                                                 & 0.4796$\pm$0.0084 & 0.4367$\pm$0.0094 \\
                           &      PINE               &                                                                 & 0.4628$\pm$0.0084 & 0.4466$\pm$0.0096 \\\cmidrule{2-5}
                           &      LongRAG            & \textbf{0.9253$\pm$0.0050}                                      & 0.4744$\pm$0.0084 & 0.4329$\pm$0.0093 \\
                           &      Hybrid             & 0.8114$\pm$0.0071                                               & 0.4803$\pm$0.0088 & \underline{0.4621$\pm$0.0092}\\
                           &  \cellcolor{gray!15}   \textbf{UMG-RAG (Ours)}      &\cellcolor{gray!15}0.7949$\pm$0.0077&\cellcolor{gray!15}\underline{0.4807$\pm$0.0082}&\cellcolor{gray!15}0.4582$\pm$0.0093\\
                           &  \cellcolor{gray!15}   \textbf{UMGP-RAG (Ours)}     &\cellcolor{gray!15}\underline{0.8745$\pm$0.0064}&\cellcolor{gray!15}\textbf{0.4961$\pm$0.0085}&\cellcolor{gray!15}\textbf{0.4748$\pm$0.0091}\\ \bottomrule
\end{tabular}
\caption{
\textbf{Results on the Natural Questions dataset (Qwen2.5-3B-Instruct).}
The best retrieval and generation scores within each retriever block are highlighted in \textbf{bold}, and the second best scores are \underline{underlined}.
}
\label{tab:nq_result_qwen}
\end{table*}
\begin{table*}[t]
\fontsize{9.5pt}{11pt}\selectfont
\centering
\begin{tabular}{c|c|c|cc}
\toprule
\multirow{2}{*}{Retriever} & \multirow{2}{*}{Method} & \multirow{2}{*}{\begin{tabular}[c]{@{}c@{}}Retrieval\\ AR@5 $\uparrow$\end{tabular}} & \multicolumn{2}{c}{Llama-3.2-3B-Instruct} \\ \cmidrule{4-5}
                           &                         &                                                                 & F1 $\uparrow$         & AR $\uparrow$        \\ \midrule
SPLADEv3                   &         RAG             & 0.7994$\pm$0.0074                                               & 0.4923$\pm$0.0081 & 0.4741$\pm$0.0097 \\ \midrule
\multirow{10}{*}{BERT}     &        RAG              & \multirow{6}{*}{0.4477$\pm$0.0090}                              & 0.2414$\pm$0.0074 & 0.2178$\pm$0.0076 \\
                           &      Summarized RAG     &                                                                 & 0.1972$\pm$0.0068 & 0.1703$\pm$0.0071 \\
                           &      LongLLMLingua      &                                                                 & 0.2573$\pm$0.0079 & 0.2429$\pm$0.0081 \\
                           &      MS-PoE             &                                                                 & 0.2435$\pm$0.0073 & 0.2192$\pm$0.0076 \\
                           &      REPLUG             &                                                                 & 0.2254$\pm$0.0071 & 0.1896$\pm$0.0071 \\
                           &      PINE               &                                                                 & 0.2084$\pm$0.0067 & 0.2097$\pm$0.0074 \\\cmidrule{2-5}
                           &      LongRAG            & 0.4441$\pm$0.0093                                               & 0.2318$\pm$0.0073 & 0.2094$\pm$0.0077 \\
                           &      Hybrid             & \underline{0.7011$\pm$0.0086}                                   & 0.4504$\pm$0.0083 & 0.4293$\pm$0.0092 \\
                           &  \cellcolor{gray!15}   \textbf{UMG-RAG (Ours)}      &\cellcolor{gray!15} 0.6852$\pm$0.0088&\cellcolor{gray!15}\underline{0.4514$\pm$0.0080}&\cellcolor{gray!15}\underline{0.4304$\pm$0.0097}\\
                           &\cellcolor{gray!15}      \textbf{UMGP-RAG (Ours)}     &\cellcolor{gray!15}\textbf{0.7797$\pm$0.0078}&\cellcolor{gray!15}\textbf{0.4811$\pm$0.0086}&\cellcolor{gray!15}\textbf{0.4533$\pm$0.0093}\\ \midrule
\multirow{10}{*}{BGE-M3}   &         RAG             & \multirow{6}{*}{0.8562$\pm$0.0068}                              & 0.4941$\pm$0.0082 & 0.4593$\pm$0.0094 \\
                           &      Summarized RAG     &                                                                 & 0.3520$\pm$0.0081 & 0.3127$\pm$0.0086 \\
                           &      LongLLMLingua      &                                                                 & 0.4631$\pm$0.0083 & 0.4462$\pm$0.0096 \\
                           &      MS-PoE             &                                                                 & 0.5005$\pm$0.0086 & 0.4709$\pm$0.0091 \\
                           &      REPLUG             &                                                                 & 0.4962$\pm$0.0085 & 0.4459$\pm$0.0089 \\
                           &      PINE               &                                                                 & 0.4452$\pm$0.0080 & 0.4603$\pm$0.0093 \\\cmidrule{2-5}
                           &      LongRAG            & \textbf{0.9101$\pm$0.0055}                                      & \underline{0.5113$\pm$0.0081}& \textbf{0.4910$\pm$0.0095}\\
                           &      Hybrid             & 0.8241$\pm$0.0071                                               & 0.4986$\pm$0.0083 & 0.4766$\pm$0.0097 \\
                           &    \cellcolor{gray!15} \textbf{UMG-RAG (Ours)}      &\cellcolor{gray!15}0.8023$\pm$0.0074&\cellcolor{gray!15}0.4873$\pm$0.0078&\cellcolor{gray!15}0.4716$\pm$0.0095\\
                           &    \cellcolor{gray!15} \textbf{UMGP-RAG (Ours)}     &\cellcolor{gray!15}\underline{0.8759$\pm$0.0063}&\cellcolor{gray!15}\textbf{0.5128$\pm$0.0084}&\cellcolor{gray!15}\underline{0.4878$\pm$0.0094}\\ \midrule
\multirow{10}{*}{\begin{tabular}[c]{@{}c@{}}Qwen3-\\ Embedding-4B\end{tabular}} &       RAG              &  \multirow{6}{*}{0.8601$\pm$0.0067}                             & 0.4939$\pm$0.0084 & 0.4621$\pm$0.0093 \\
                           &      Summarized RAG     &                                                                 & 0.3537$\pm$0.0083 & 0.3162$\pm$0.0087 \\
                           &      LongLLMLingua      &                                                                 & 0.4561$\pm$0.0083 & 0.4346$\pm$0.0098 \\
                           &      MS-PoE             &                                                                 & 0.5046$\pm$0.0083 & 0.4720$\pm$0.0094 \\
                           &      REPLUG             &                                                                 & 0.5120$\pm$0.0084 & 0.4642$\pm$0.0088 \\
                           &      PINE               &                                                                 & 0.4291$\pm$0.0082 & 0.4438$\pm$0.0095 \\\cmidrule{2-5}
                           &      LongRAG            & \textbf{0.9253$\pm$0.0050}                                      & \underline{0.5145$\pm$0.0082}& \underline{0.4949$\pm$0.0096}\\
                           &      Hybrid             & 0.8114$\pm$0.0071                                               & 0.4916$\pm$0.0083 & 0.4741$\pm$0.0096 \\
                           &  \cellcolor{gray!15}   \textbf{UMG-RAG (Ours)}      &\cellcolor{gray!15}0.7949$\pm$0.0077&\cellcolor{gray!15}0.4974$\pm$0.0080&\cellcolor{gray!15}0.4790$\pm$0.0093\\
                           &  \cellcolor{gray!15}   \textbf{UMGP-RAG (Ours)}     &\cellcolor{gray!15}\underline{0.8745$\pm$0.0064}&\cellcolor{gray!15}\textbf{0.5220$\pm$0.0084}&\cellcolor{gray!15}\textbf{0.4984$\pm$0.0094}\\ \bottomrule
\end{tabular}
\caption{
\textbf{Results on the Natural Questions dataset (Llama-3.2-3B-Instruct).}
The best retrieval and generation scores within each retriever block are highlighted in \textbf{bold}, and the second best scores are \underline{underlined}.
}
\label{tab:nq_result_llama}
\end{table*}

\begin{table*}[t]
\fontsize{9.5pt}{11pt}\selectfont
\centering
\begin{tabular}{c|c|c|cc}
\toprule
\multirow{2}{*}{Retriever} & \multirow{2}{*}{Method} & \multirow{2}{*}{\begin{tabular}[c]{@{}c@{}}Retrieval\\ AR@5 $\uparrow$\end{tabular}} & \multicolumn{2}{c}{Qwen2.5-3B-Instruct} \\ \cmidrule{4-5}
                           &                         &                                                                 & F1 $\uparrow$         & AR $\uparrow$        \\ \midrule
SPLADEv3                   &         RAG             & 0.6043$\pm$0.0065                                               & 0.4613$\pm$0.0061 & 0.3850$\pm$0.0063 \\ \midrule
\multirow{10}{*}{BERT}     &        RAG              & \multirow{6}{*}{0.3638$\pm$0.0067}                              & 0.2599$\pm$0.0055 & 0.2087$\pm$0.0053 \\
                           &      Summarized RAG     &                                                                 & 0.2275$\pm$0.0053 & 0.1864$\pm$0.0054 \\
                           &      LongLLMLingua      &                                                                 & 0.2833$\pm$0.0055 & 0.2271$\pm$0.0056 \\
                           &      MS-PoE             &                                                                 & 0.2612$\pm$0.0054 & 0.2112$\pm$0.0054 \\
                           &      REPLUG             &                                                                 & 0.1798$\pm$0.0049 & 0.1466$\pm$0.0047 \\
                           &      PINE               &                                                                 & 0.2424$\pm$0.0052 & 0.2039$\pm$0.0053 \\\cmidrule{2-5}
                           &      LongRAG            & 0.3230$\pm$0.0062                                               & 0.2299$\pm$0.0051 & 0.1861$\pm$0.0053 \\
                           &      Hybrid             & \underline{0.5207$\pm$0.0066}                                   & 0.4081$\pm$0.0062 & 0.3311$\pm$0.0060 \\
                           &    \cellcolor{gray!15} \textbf{UMG-RAG (Ours)}      &\cellcolor{gray!15}0.5134$\pm$0.0068&\cellcolor{gray!15}\underline{0.4163$\pm$0.0062}&\cellcolor{gray!15}\underline{0.3416$\pm$0.0064}\\
                           &     \cellcolor{gray!15}\textbf{UMGP-RAG (Ours)}     &\cellcolor{gray!15}\textbf{0.5830$\pm$0.0067}&\cellcolor{gray!15}\textbf{0.4453$\pm$0.0061}&\cellcolor{gray!15}\textbf{0.3645$\pm$0.0065}\\ \midrule
\multirow{10}{*}{BGE-M3}   &         RAG             & \multirow{6}{*}{\textbf{0.7811$\pm$0.0057}}                     & 0.4613$\pm$0.0064 & 0.3798$\pm$0.0065 \\
                           &      Summarized RAG     &                                                                 & 0.3856$\pm$0.0058 & 0.3161$\pm$0.0064 \\
                           &      LongLLMLingua      &                                                                 & 0.4706$\pm$0.0061 & 0.3807$\pm$0.0065 \\
                           &      MS-PoE             &                                                                 & 0.4622$\pm$0.0059 & 0.3805$\pm$0.0066 \\
                           &      REPLUG             &                                                                 & 0.3941$\pm$0.0060 & 0.3150$\pm$0.0061 \\
                           &      PINE               &                                                                 & 0.4442$\pm$0.0062 & 0.3787$\pm$0.0065 \\\cmidrule{2-5}
                           &      LongRAG            & \underline{0.7457$\pm$0.0059}                                   & 0.4658$\pm$0.0061 & 0.3845$\pm$0.0065 \\
                           &      Hybrid             & 0.6471$\pm$0.0063                                               & \underline{0.4775$\pm$0.0063}& \underline{0.3975$\pm$0.0063}\\
                           &    \cellcolor{gray!15} \textbf{UMG-RAG (Ours)}      &\cellcolor{gray!15}0.6273$\pm$0.0064&\cellcolor{gray!15}0.4749$\pm$0.0062&\cellcolor{gray!15}0.3957$\pm$0.0066\\
                           &     \cellcolor{gray!15}\textbf{UMGP-RAG (Ours)}     &\cellcolor{gray!15}0.7002$\pm$0.0061&\cellcolor{gray!15}\textbf{0.5034$\pm$0.0062}&\cellcolor{gray!15}\textbf{0.4216$\pm$0.0068}\\ \midrule
\multirow{10}{*}{\begin{tabular}[c]{@{}c@{}}Qwen3-\\ Embedding-4B\end{tabular}} &       RAG              &  \multirow{6}{*}{\textbf{0.7971$\pm$0.0053}}                    & 0.4883$\pm$0.0060 & 0.4041$\pm$0.0065 \\
                           &      Summarized RAG     &                                                                 & 0.3976$\pm$0.0058 & 0.3261$\pm$0.0063 \\
                           &      LongLLMLingua      &                                                                 & 0.4603$\pm$0.0061 & 0.3693$\pm$0.0063 \\
                           &      MS-PoE             &                                                                 & \textbf{0.4948$\pm$0.0060}& \textbf{0.4073$\pm$0.0067}\\
                           &      REPLUG             &                                                                 & 0.4075$\pm$0.0060 & 0.3252$\pm$0.0063 \\
                           &      PINE               &                                                                 & 0.4621$\pm$0.0061 & 0.3980$\pm$0.0065 \\\cmidrule{2-5}
                           &      LongRAG            & \underline{0.7280$\pm$0.0062}                                   & 0.4524$\pm$0.0061 & 0.3748$\pm$0.0066 \\
                           &      Hybrid             & 0.6091$\pm$0.0063                                               & 0.4626$\pm$0.0060 & 0.3837$\pm$0.0062 \\
                           &    \cellcolor{gray!15} \textbf{UMG-RAG (Ours)}      &\cellcolor{gray!15}0.5845$\pm$0.0068&\cellcolor{gray!15}0.4622$\pm$0.0060&\cellcolor{gray!15}0.3820$\pm$0.0064\\
                           &     \cellcolor{gray!15}\textbf{UMGP-RAG (Ours)}     &\cellcolor{gray!15}0.6761$\pm$0.0064&\cellcolor{gray!15}\underline{0.4919$\pm$0.0060}&\cellcolor{gray!15}\underline{0.4071$\pm$0.0066}\\ \bottomrule
\end{tabular}
\caption{
\textbf{Results on the HotPotQA dataset (Qwen2.5-3B-Instruct).}
The best retrieval and generation scores within each retriever block are highlighted in \textbf{bold}, and the second best scores are \underline{underlined}.
}
\label{tab:hotpotqa_result_qwen}
\end{table*}
\begin{table*}[t]
\fontsize{9.5pt}{11pt}\selectfont
\centering
\begin{tabular}{c|c|c|cc}
\toprule
\multirow{2}{*}{Retriever} & \multirow{2}{*}{Method} & \multirow{2}{*}{\begin{tabular}[c]{@{}c@{}}Retrieval\\ AR@5 $\uparrow$\end{tabular}} & \multicolumn{2}{c}{Llama-3.2-3B-Instruct} \\ \cmidrule{4-5}
                           &                         &                                                                 & F1 $\uparrow$         & AR $\uparrow$        \\ \midrule
SPLADEv3                   &         RAG             & 0.6043$\pm$0.0065                                               & 0.4798$\pm$0.0062 & 0.3966$\pm$0.0069 \\ \midrule
\multirow{10}{*}{BERT}     &        RAG              & \multirow{6}{*}{0.3638$\pm$0.0067}                              & 0.2803$\pm$0.0055 & 0.2279$\pm$0.0057 \\
                           &      Summarized RAG     &                                                                 & 0.2469$\pm$0.0055 & 0.2025$\pm$0.0053 \\
                           &      LongLLMLingua      &                                                                 & 0.3021$\pm$0.0057 & 0.2362$\pm$0.0057 \\
                           &      MS-PoE             &                                                                 & 0.2860$\pm$0.0057 & 0.2339$\pm$0.0056 \\
                           &      REPLUG             &                                                                 & 0.2560$\pm$0.0053 & 0.1984$\pm$0.0052 \\
                           &      PINE               &                                                                 & 0.2557$\pm$0.0052 & 0.2180$\pm$0.0055 \\\cmidrule{2-5}
                           &      LongRAG            & 0.3230$\pm$0.0062                                               & 0.2625$\pm$0.0056 & 0.2104$\pm$0.0056 \\
                           &      Hybrid             & \underline{0.5207$\pm$0.0066}                                   & \underline{0.4245$\pm$0.0059}& \underline{0.3479$\pm$0.0064}\\
                           &    \cellcolor{gray!15} \textbf{UMG-RAG (Ours)}      &\cellcolor{gray!15}0.5134$\pm$0.0068&\cellcolor{gray!15}0.4223$\pm$0.0060&\cellcolor{gray!15}0.3470$\pm$0.0067\\
                           &     \cellcolor{gray!15}\textbf{UMGP-RAG (Ours)}     &\cellcolor{gray!15}\textbf{0.5830$\pm$0.0067}&\cellcolor{gray!15}\textbf{0.4460$\pm$0.0059}&\cellcolor{gray!15}\textbf{0.3655$\pm$0.0063}\\ \midrule
\multirow{10}{*}{BGE-M3}   &         RAG             & \multirow{6}{*}{\textbf{0.7811$\pm$0.0057}}                     & 0.4913$\pm$0.0061 & 0.4048$\pm$0.0064 \\
                           &      Summarized RAG     &                                                                 & 0.4109$\pm$0.0060 & 0.3366$\pm$0.0061 \\
                           &      LongLLMLingua      &                                                                 & 0.4668$\pm$0.0062 & 0.3771$\pm$0.0062 \\
                           &      MS-PoE             &                                                                 & \underline{0.5001$\pm$0.0062}& 0.4123$\pm$0.0068 \\
                           &      REPLUG             &                                                                 & 0.4287$\pm$0.0060 & 0.3362$\pm$0.0065 \\
                           &      PINE               &                                                                 & 0.4409$\pm$0.0060 & 0.3941$\pm$0.0065 \\\cmidrule{2-5}
                           &      LongRAG            & \underline{0.7457$\pm$0.0059}                                   & 0.4799$\pm$0.0059 & 0.3984$\pm$0.0066 \\
                           &      Hybrid             & 0.6471$\pm$0.0063                                               & 0.4986$\pm$0.0060 & \underline{0.4175$\pm$0.0069}\\
                           &    \cellcolor{gray!15} \textbf{UMG-RAG (Ours)}      &\cellcolor{gray!15}0.6273$\pm$0.0064&\cellcolor{gray!15}0.4827$\pm$0.0059&\cellcolor{gray!15}0.4018$\pm$0.0068\\
                           &     \cellcolor{gray!15}\textbf{UMGP-RAG (Ours)}     &\cellcolor{gray!15}0.7002$\pm$0.0061&\cellcolor{gray!15}\textbf{0.5133$\pm$0.0060}&\cellcolor{gray!15}\textbf{0.4275$\pm$0.0066}\\ \midrule
\multirow{10}{*}{\begin{tabular}[c]{@{}c@{}}Qwen3-\\ Embedding-4B\end{tabular}} &       RAG              &  \multirow{6}{*}{\textbf{0.7971$\pm$0.0053}}                    & 0.4991$\pm$0.0060 & 0.4132$\pm$0.0065 \\
                           &      Summarized RAG     &                                                                 & 0.4315$\pm$0.0060 & 0.3546$\pm$0.0061 \\
                           &      LongLLMLingua      &                                                                 & 0.4644$\pm$0.0060 & 0.3738$\pm$0.0063 \\
                           &      MS-PoE             &                                                                 & \textbf{0.5172$\pm$0.0059}& \textbf{0.4295$\pm$0.0065}\\
                           &      REPLUG             &                                                                 & 0.4343$\pm$0.0061 & 0.3411$\pm$0.0064 \\
                           &      PINE               &                                                                 & 0.4463$\pm$0.0059 & 0.4027$\pm$0.0067 \\\cmidrule{2-5}
                           &      LongRAG            & \underline{0.7280$\pm$0.0062}                                   & 0.4765$\pm$0.0061 & 0.3961$\pm$0.0063 \\
                           &      Hybrid             & 0.6091$\pm$0.0063                                               & 0.4709$\pm$0.0060 & 0.3904$\pm$0.0066 \\
                           &    \cellcolor{gray!15} \textbf{UMG-RAG (Ours)}      &\cellcolor{gray!15}0.5845$\pm$0.0068&\cellcolor{gray!15}0.4752$\pm$0.0060&\cellcolor{gray!15}0.3929$\pm$0.0068\\
                           &     \cellcolor{gray!15}\textbf{UMGP-RAG (Ours)}     &\cellcolor{gray!15}0.6761$\pm$0.0064&\cellcolor{gray!15}\underline{0.5068$\pm$0.0059}&\cellcolor{gray!15}\underline{0.4200$\pm$0.0064}\\ \bottomrule
\end{tabular}
\caption{
\textbf{Results on the HotPotQA dataset (Llama-3.2-3B-Instruct).}
The best retrieval and generation scores within each retriever block are highlighted in \textbf{bold}, and the second best scores are \underline{underlined}.
}
\label{tab:hotpotqa_result_llama}
\end{table*}

\section{AI Assistant Acknowledgement}

AI assistance was used to support code debugging and improve the clarity of the manuscript. The authors carefully reviewed, edited, and verified all AI-assisted content and take full responsibility for the final version of this publication.

\section{Computational Infrastructure and Budget}
All experiments were conducted on a single NVIDIA RTX A5000 GPU with 24GB memory. Because the proposed methods are training-free and do not update retriever or generator parameters, the computational budget is reported as retrieval preprocessing time, generation time, and peak memory usage per question rather than training GPU hours. The detailed time and memory measurements are reported in Appendix \ref{sec:cost}.
\end{document}